\documentclass{article}

\usepackage[preprint]{neurips_2026}

\usepackage[utf8]{inputenc} 
\usepackage[T1]{fontenc}    
\usepackage{hyperref}       
\usepackage{url}            
\usepackage{booktabs}       
\usepackage{amsfonts}       
\usepackage{nicefrac}       
\usepackage{microtype}      
\usepackage{xcolor}         

\usepackage{amssymb} 
\usepackage{multirow}
\usepackage{pifont}
\newcommand{\xmark}{\ding{55}}
\usepackage{xcolor}
\usepackage{subcaption} 
\usepackage{graphicx}
\usepackage{amsmath}
\usepackage{amsthm}
\usepackage{booktabs}
\usepackage{algorithm}
\usepackage{algorithmic}

\title{	
Congestion-Aware Dynamic Axonal Delay for Spiking Neural Networks}

\author{%
  Dewei Bai\textsuperscript{*}
  \And
  Hongxiang Peng\textsuperscript{*}
  \And
  Yunyun Zeng
  \And
  Ziyu Zhang
  \And
  Hong Qu\textsuperscript{\#} \\
  University of Electronic Science and Technology of China \\
  \textsuperscript{*}Equal contribution.
  \quad
  \textsuperscript{\#}Corresponding author.
}

\begin{document}

\maketitle

\begin{abstract}
  Spiking Neural Networks (SNNs) are widely regarded as an energy-efficient paradigm for modeling and processing temporal and event-driven information. Incorporating delays in SNNs has been proven to be an effective mechanism for improving spike alignment in event-driven tasks. However, existing delay learning approaches predominantly assign static delays to individual synapses, resulting in a large number of delay parameters and limited adaptability to input-dependent activity dynamics. To this end, we propose a Congestion-Aware Dynamic Axonal Delay (CADAD) mechanism, which decomposes the delay into a channel-wise static base delay for temporal structuring and a global, activity-conditioned shift that dynamically regulates the state update rate under varying spike intensities. The delay parameters are learned using differentiable linear interpolation and discretized at inference time, preserving the benefits of dynamic delay modulation while incurring only minimal additional cost. Experiments on speech benchmarks, including the Spiking Heidelberg Dataset, Spiking Speech Commands, and Google Speech Commands, demonstrate that introducing congestion-aware delays into synaptic signal transmission effectively improves accuracy on temporal tasks, notably achieving 93.75\% accuracy on SHD, 80.69\% accuracy on SSC, and 95.58\% on GSC-35, while reducing the parameter count by approximately 50\% compared to state-of-the-art delay-based methods with the same architecture.
\end{abstract}

\section{Introduction} 
Owing to their event-driven computational paradigm, Spiking Neural Networks (SNNs) exhibit significant advantages in low-power and energy-efficient intelligent computing. In such networks, information propagates across space and time via discrete spikes, with the representational capacity and training stability being highly dependent on the temporal alignment, propagation, and integration of spike signals.

From the perspective of computational neuroscience, spiking neurons can be regarded as time-sensitive coincidence detectors \cite{Konig1996,Rossant2011}, where effective information processing relies on the relative timing of spike arrivals rather than the firing rate of a single neuron. Signal delays along axons and synapses directly shape these arrival times, and the heterogeneity of such delays provides critical degrees of freedom for modeling complex spatiotemporal patterns \cite{Izhikevich2006}. Extensive biological research further indicates that synaptic and transmission delays are not static; instead, they can be regulated through learning processes, suggesting that temporal regulation itself is a fundamental component of neural computation \cite{Bowers2017a}.

Inspired by these findings, recent studies have introduced learnable transmission delays into SNNs to explicitly enhance temporal modeling capabilities, achieving performance gains across various time-sensitive tasks. However, most existing methods model delays as static connection-level parameters, where each synapse corresponds to a fixed temporal offset that remains unchanged after training. Such modeling implicitly assumes a stable temporal alignment between pre- and post-synaptic activities—an assumption that often falters in deep networks or scenarios with highly non-stationary spiking activity. In practice, the significance of spike timing and the optimal alignment typically depend on the current global state of the network rather than a fixed structural configuration.

\begin{figure*}[t]
  \centering
  \begin{subfigure}[b]{0.32\textwidth}
    \centering
    \includegraphics[width=\linewidth]{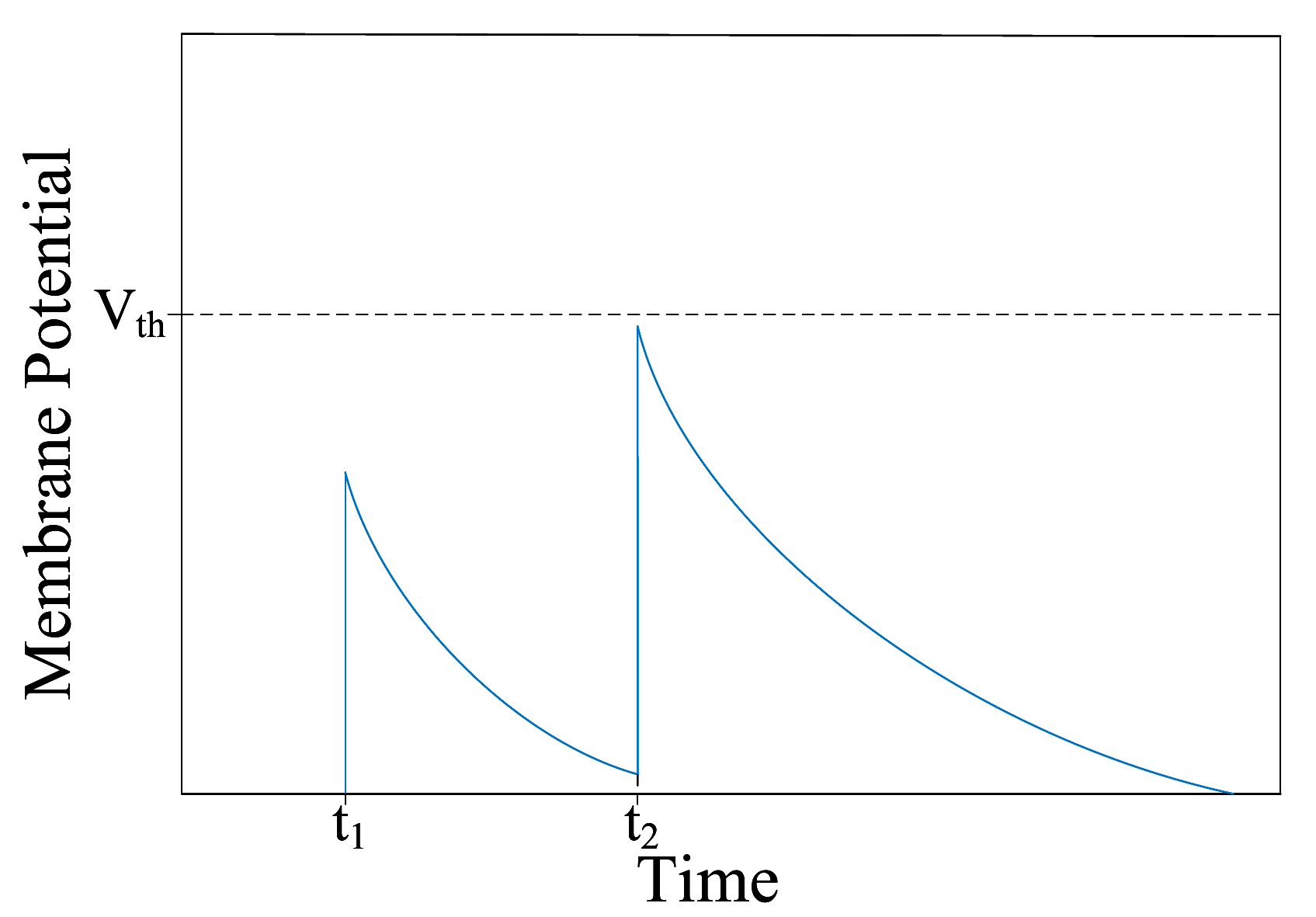}
    \caption{No delay}
    \label{fig:concept_nodelay}
  \end{subfigure}
  \hfill
  \begin{subfigure}[b]{0.32\textwidth}
    \centering
    \includegraphics[width=\linewidth]{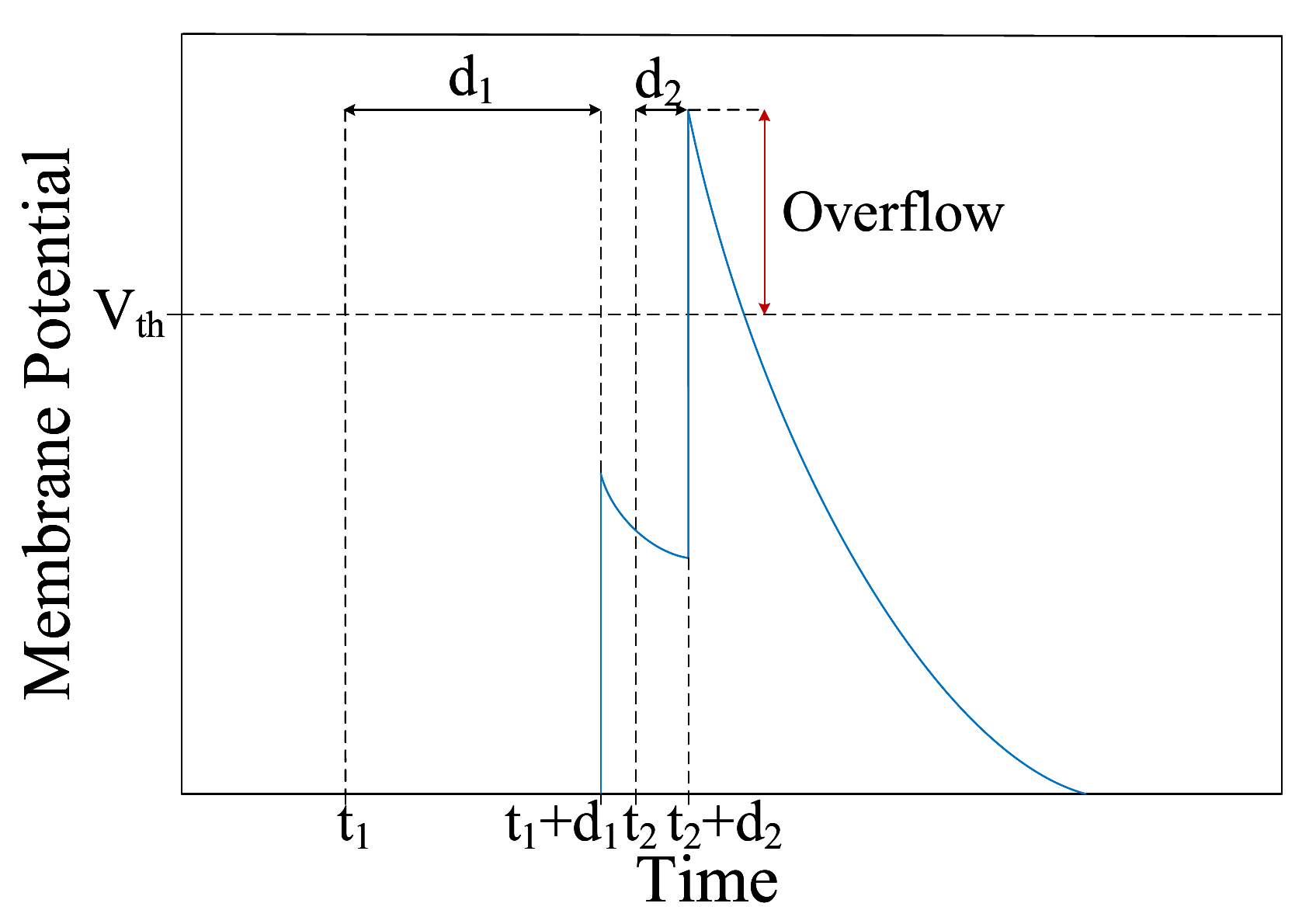}
    \caption{Static delay}
    \label{fig:concept_static}
  \end{subfigure}
  \hfill
  \begin{subfigure}[b]{0.32\textwidth}
    \centering
    \includegraphics[width=\linewidth]{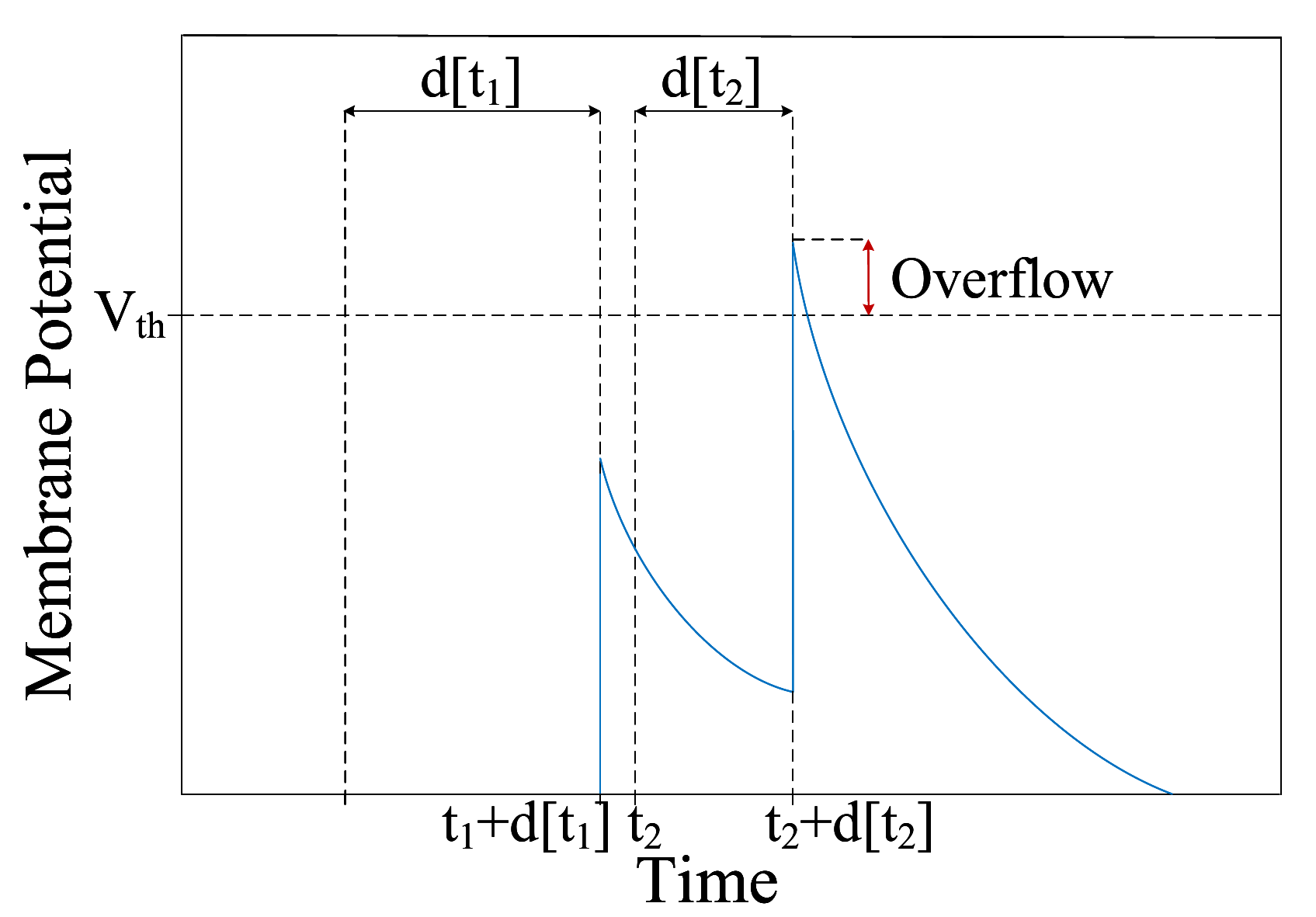}
    \caption{Dynamic delay}
    \label{fig:concept_dynamic}
  \end{subfigure}

  \caption{Conceptual comparison of delay mechanisms. (a) Without delays, spikes arrive at $t_1$ and $t_2$ asynchronously, causing information loss due to membrane leakage. (b) Static delays add fixed time intervals, improving integration but resulting in misalignment and broad wasteful overflow (red arrow) when input timing jitters. (c) Dynamic delays adaptively align inputs to a target moment, producing a decisive, high-precision firing event with minimized information loss.}
  \label{fig:delay_mechanism_comparison}
\end{figure*}

Based on these observations, we argue that temporal alignment should be treated as a dynamic, signal-level modulation process rather than an immutable structural property. Along this line, this paper focuses on the problem of adaptive temporal alignment under dynamic and non-stationary spiking conditions and proposes a \textbf{Congestion-Aware Dynamic Axonal Delay} (CADAD) mechanism. As conceptually illustrated in Figure~\ref{fig:delay_mechanism_comparison}, unlike static delays which often lead to misalignment and overflow when input timing jitters, our dynamic mechanism adaptively aligns input spikes to a target moment, thereby minimizing quantization loss and ensuring decisive firing events. This approach decomposes signal transmission delays into channel-wise static base components and dynamic offsets modulated by the global neural activity state. This factorization allows the network to adaptively adjust state update rates under varying spike load intensities, thereby enhancing dynamic stability while preserving event-level temporal semantics.

Our contributions are summarized as follows:
\begin{enumerate}
    \item We propose a congestion-aware dynamic axonal delay mechanism that enables neural-activity-dependent temporal modulation, breaking the limitations of existing static connection-level delay modeling.
    \item We design an efficient and fully differentiable linear interpolation approximation for delays, enabling stable end-to-end optimization of time-varying delays with low computational cost.
    \item Experimental results on multiple speech benchmarks demonstrate that our method achieves competitive accuracy against state-of-the-art delay-based SNN approaches with the same architecture while significantly reducing the number of delay parameters.
\end{enumerate}

\section{Related work}

\subsection{Training deep spiking neural networks}

Training deep spiking neural networks (SNNs) has long been considered challenging due to the non-differentiability of spike events and the complex temporal dynamics induced by recurrent membrane states. Surrogate gradient methods~\cite{neftci2018surrogate,slayer} address this issue by replacing the discontinuous spike function with smooth approximations during backpropagation, enabling end-to-end gradient-based optimization. In parallel, ANN-to-SNN conversion approaches~\cite{ann2snn_1,ann2snn_2,ann2snn_3} leverage pretrained artificial neural networks to initialize SNNs, significantly improving training stability and performance while minimizing conversion error.

Beyond training algorithms, several works have focused on improving the expressivity and optimization behavior of spiking neurons. The Parametric Leaky Integrate-and-Fire (PLIF) model~\cite{plif} introduces learnable membrane time constants, while adaptive threshold mechanisms~\cite{bellec2018,hammouamri2022mitigating} dynamically regulate firing activity to improve stability and continual learning performance. Architectural advances such as Spike-Element-Wise ResNet~\cite{sew} mitigate gradient vanishing and explosion in deep SNNs, enabling successful training of networks with over one hundred layers. More recently, transformer-inspired spiking architectures~\cite{spikformer,spikegpt,spikingformer} adapt attention mechanisms to the spiking domain, further narrowing the performance gap between SNNs and conventional neural networks.

Although prior work has substantially improved the trainability and performance of deep SNNs through advances in training strategies, neuron dynamics, and architectural design, temporal modeling remains largely implicit, relying on membrane dynamics or fixed structural constraints. Such indirect representations often limit fine-grained temporal alignment in event-driven tasks with complex temporal dependencies and highly non-stationary spike activity. This highlights the need for lightweight and explicit temporal modulation mechanisms that preserve training stability.


\subsection{Learning delays in spiking neural networks}

Synaptic delays serve as a fundamental mechanism for temporal information processing in both biological and theoretical neural systems. Early theoretical studies established that SNNs equipped with adjustable delays possess strictly superior functional representation capabilities compared to threshold circuits relying solely on weight adaptation \cite{Maass1999}. This foundational insight has catalyzed the development of various delay learning methodologies.

Initial efforts in delay learning typically treated each synaptic delay as a static hyperparameter. For instance, learnable temporal offsets were introduced through dilated convolutions, where the positions within the convolutional kernels are optimized to achieve specific time shifts \cite{hammouamri2024learning}. As the understanding of delays deepened, researchers began integrating delay mechanisms into more sophisticated architectures to exploit their potential in structured tasks. Notably, Delay-DSGN incorporates learnable delays into Spiking Graph Neural Networks to adaptively regulate the propagation of historical information, thereby enhancing the modeling of dynamic graph evolution \cite{wangDelayDSGNDynamicSpiking}. In temporal reasoning, combining delays with temporal encoding has been shown to strengthen the capture of complex relational and sequential dependencies \cite{xiaoTemporalSpikingNeural2024}. Furthermore, the DelRec framework emphasizes the necessity of differentiable delays for modeling long-term temporal dependencies within recurrent SNN structures, achieving state-of-the-art performance across diverse sequence tasks \cite{queantDelRecLearningDelays2025}.

Parallel to architectural innovations, joint learning frameworks have been explored to improve biological plausibility and temporal adaptability. Recent work has proposed the co-optimization of synaptic delays, connection weights, and adaptive neuronal parameters, allowing the network to learn rich dynamic patterns through coordinated parameter updates \cite{deckersColearningSynapticDelays2024}. Biologically inspired local update rules, such as extensions of Spike-Timing-Dependent Plasticity (STDP) to the delay domain \cite{dominijanniExtendingSpikeTimingDependent2025} and three-factor learning rules for online adaptation \cite{vassalloThreeFactorDelay2026}, have also been developed.
  
Despite these advancements, most existing methods still model delays as static parameters or fixed structural attributes, which limits their adaptability in non-stationary spiking environments and often incurs substantial optimization overhead. To address these limitations, we propose a congestion-aware dynamic axonal delay mechanism. By decomposing the transmission delay into a structurally stable component and an activity-dependent dynamic modulation term, our approach achieves adaptive temporal alignment under complex spatiotemporal dynamics while significantly reducing the parameter footprint.

\section{Methods}
\subsection{Delay in spiking neuron model}
We consider axonal transmission delays within a standard Leaky Integrate-and-Fire (LIF) neuron model.
Let $v(t)$ denote the membrane potential and $I(t)$ the synaptic input current.
The subthreshold dynamics are governed by a first-order low-pass filter:
\begin{equation}
\tau \frac{dv(t)}{dt} = -\,v(t) + I(t),
\label{eq:lif_cont}
\end{equation}
where $\tau>0$ is the membrane time constant. When $v(t)$ reaches the threshold
$V_{\mathrm{th}}$, the neuron emits a spike and the membrane is reset to $V_{\mathrm{reset}}$.

For discrete-time simulation with step $\Delta t$, the membrane update can be written with an effective leak factor $\lambda\in(0,1)$:
\begin{equation}
\tilde{v}[t] = \lambda\,v[t-1] + I[t],
\label{eq:lif_tmp}
\end{equation}
\begin{equation}
S[t] = \Theta\!\big(\tilde{v}[t] - V_{\mathrm{th}}\big), \qquad
\Theta(u)=
\begin{cases}
1,& u\ge 0,\\
0,& u<0,
\end{cases}
\label{eq:lif_spk}
\end{equation}
\begin{equation}
v[t] = \tilde{v}[t]\,(1-S[t]) + V_{\mathrm{reset}}\,S[t],
\label{eq:lif_reset}
\end{equation}
where $I[t]$ is the input drive at time step $t$, 
$S[t]\!\in\!\{0,1\}$ is the spike output, and $\tilde{v}[t]$ is the pre-spike potential.
Eqs.~\eqref{eq:lif_tmp}--\eqref{eq:lif_reset} jointly realize the leaky integration, 
threshold firing, and reset process of the LIF neuron.

We model axonal transmission delays from neuron $j$ in layer $l\!-\!1$ to layer $l$, where each synaptic connection has a weight $w_{ij}^{l}$, while each presynaptic neuron is associated with a single effective axonal delay $d_{j}^{l}$.
This abstraction keeps one effective transmission delay between each axonal output and the subsequent layer, rather than modeling multiple synapse-specific delays.

Accordingly, a feed-forward spiking neuron with delays is parameterized by the synaptic weight matrix $W = (w_{ij}^{l})$ and the axonal delay vector $D = (d_{j}^{l}) \in \mathbb{R}^{+}$. 
The synaptic input current to neuron $i$ in layer $l$ at time step $t$ is given by
\begin{equation}
    I_i^{l}[t] = \sum_{j} w_{ij}^{l} \, S_{j}^{l-1}[t - d_{j}^{l}].
\end{equation}

\begin{figure*}
    \centering
    \includegraphics[width=1\linewidth]{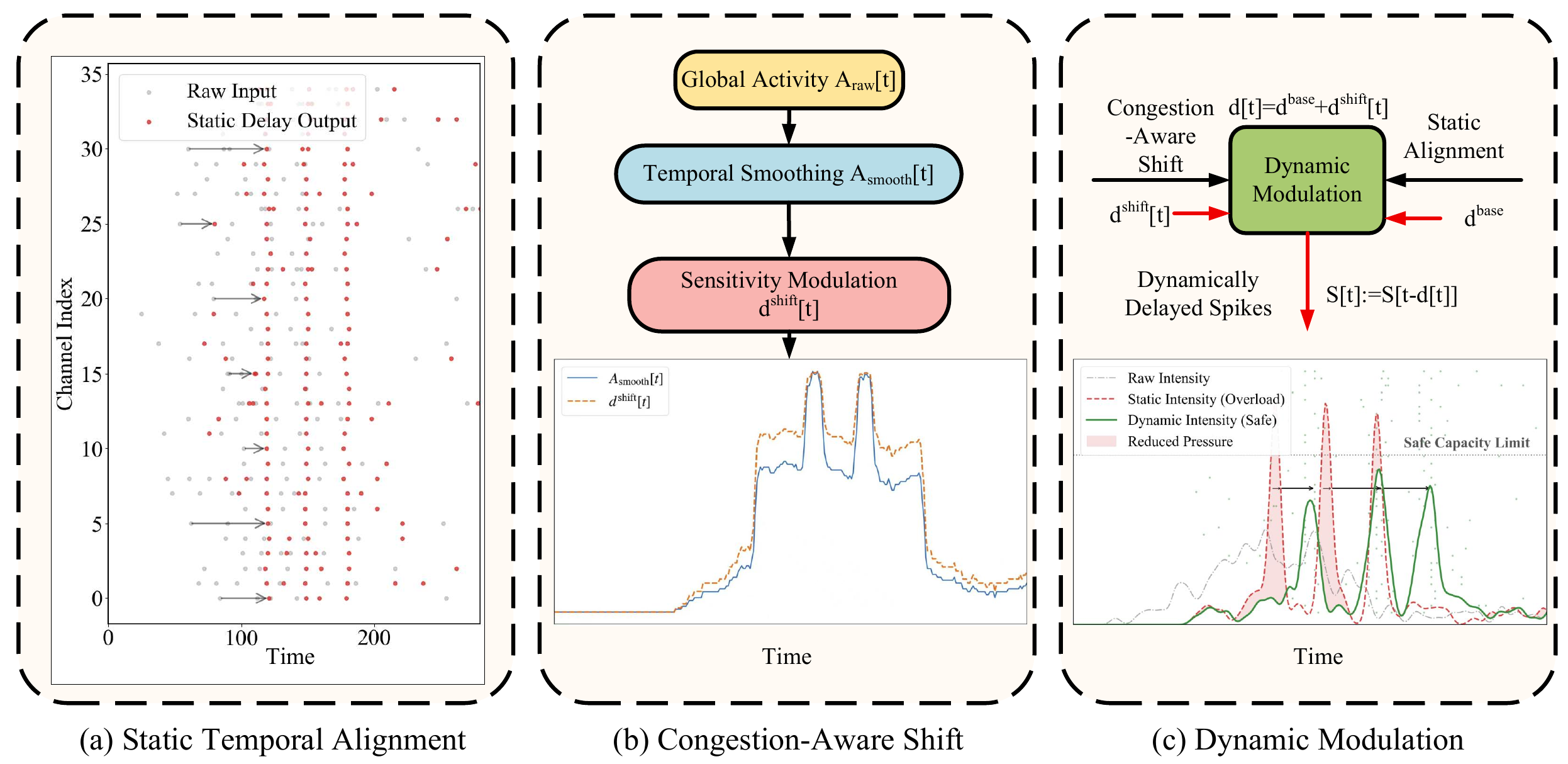}
    \caption{Overview of the congestion-aware dynamic axonal delay mechanism. (a) Traditional static delays succeed in aligning spikes but can also cause instantaneous congestion, overloading the network. (b) The dynamic shift $d^{\mathrm{shift}}[t]$ generated by the global congestion metric $A_{\mathrm{smooth}}[t]$. (c) By combining static alignment with dynamic shifts, the mechanism adaptively redistributes spikes (green curve) to mitigate excessive input concentration and reduce information loss.}
    \label{fig:overview_CADAD}
\end{figure*}

\subsection{Congestion-aware dynamic axonal delay}

In event-driven spiking networks, periods of high activity may cause a large number of event-related spikes to arrive within a narrow temporal window, giving rise to what we refer to as a \emph{high-load regime}. 
Under such conditions, a large number of state updates are triggered within a short time span, leading to abrupt membrane potential changes and excessive state update pressure. 
As a result, the discrete spike generation process becomes highly sensitive to small membrane fluctuations, which amplifies information loss during spike-based discretization and destabilizes network state transitions.

Existing delay schemes typically rely on static or heuristic parameters and are therefore unable to adapt to transient variations in activity intensity.
To address this issue, we propose the \emph{Congestion-Aware Dynamic Axonal Delay} mechanism (Figure~\ref{fig:overview_CADAD}), which adaptively modulates spike transmission latency along the axon in an activity-conditioned manner.
By dynamically regulating the temporal distribution of state updates, the proposed approach enables congestion-aware temporal modulation while preserving event-level semantics.
We treat global spike activity as an observable proxy for state update pressure and apply temporal smoothing to extract the global congestion metric $A_{\mathrm{smooth}}[t]$ to suppress high-frequency stochasticity:
\begin{equation}
    A_{\mathrm{raw}}[t] = \frac{1}{C} \sum_{j=1}^{C} S_j[t-d_{j}^{\mathrm{base}}]\in {\mathbb{R}}.
\end{equation}
Here, $\mathbf{d}^{\mathrm{base}}\in {\mathbb{R}}^{C}$ is a channel-wise static delay vector, and $d_j^{\mathrm{base}}$ denotes the base delay assigned to channel $j$.
\begin{equation}
    A_{\mathrm{smooth}}[t] = \frac{1}{k_s} \sum_{i=1}^{k_s} A_{\mathrm{raw}}[t-k_s+i],
\end{equation}
where $k_s$ is the temporal smoothing window size. The congestion-aware raw dynamic shift $\bar d^{\mathrm{shift}}[t]$ is then generated via a learnable sensitivity modulation:
\begin{equation}
    \bar d^{\mathrm{shift}}[t] = \mathcal{S}(e) \cdot D_{\max} \cdot \tanh(\gamma \cdot A_{\mathrm{smooth}}[t])\in {\mathbb{R}},
    \label{eq:tanh}
\end{equation}
where $\gamma$ is a hyperparameter controlling the sensitivity to congestion. $\mathcal{S}(e)$ is a scaling factor at epoch $e$.

We adopt an exponential annealing strategy for $\mathcal{S}(e)$ to let the network first learn the rough temporal topology and then progressively focus on fine-grained alignment:
\begin{equation}
    \mathcal{S}(e) = \max \left( S_{\max} \cdot \left( \frac{S_{\min}}{S_{\max}} \right)^{\frac{e}{E_{\mathrm{decay}}}}, S_{\min} \right)
\end{equation}
For the $E_{\mathrm{decay}}=0$ ablation, we set $\mathcal{S}(e)=S_{\min}$ from the beginning of training.

To make the time reparameterization well-defined, we apply a causal slope limiter to the dynamic shift:
\begin{equation}
    d^{\mathrm{shift}}[0] = \bar d^{\mathrm{shift}}[0], \qquad
    d^{\mathrm{shift}}[t] =
    d^{\mathrm{shift}}[t-1] +
    \operatorname{clip}\!\left(
    \bar d^{\mathrm{shift}}[t] - \bar d^{\mathrm{shift}}[t-1],
    -0.99, 0.99
    \right).
\end{equation}
This keeps adjacent dynamic-shift changes below one discrete time step.
Since the base delay is time-independent and the final clamp is non-expansive, the continuous delay used during differentiable training also satisfies $|d[t]-d[t-1]|\le 0.99$.
The purpose of this constraint is to keep high-load rising segments in the local regime $0<d'(t)<1$, while falling segments remain valid because the emission-time map is still monotone.

\begin{equation}
    d_i[t] = \operatorname{Clamp}\left( d_i^{\mathrm{base}} + d^{\mathrm{shift}}[t], \, 0, \, D_{\max} \right), \qquad i=1,\ldots,C.
\end{equation}
Here, $d_i^{\mathrm{base}}$ compensates for systematic, channel-specific latency offsets, thereby aligning event-related spikes into a shared internal temporal reference frame. In contrast, $d^{\mathrm{shift}}[t]$ is a global, time-varying shift shared across channels, which does not change event structure but modulates the load intensity of state updates. This factorization introduces an explicit inductive bias: static delays capture persistent, channel-specific temporal structure, while dynamic delays model transient, context-dependent variations in update pressure. For a fully connected layer with $N_{\mathrm{in}}$ input channels and $N_{\mathrm{out}}$ output neurons, CADAD learns one base-delay parameter per input channel, so the learnable delay parameters scale as $O(N_{\mathrm{in}})$ rather than $O(N_{\mathrm{in}}N_{\mathrm{out}})$ as in synapse-specific delay parameterizations. This separation avoids entangling structural temporal alignment with fast-varying activity dynamics, which would otherwise complicate optimization and destabilize learning.

\subsection{Theoretical interpretation}

CADAD can be interpreted as an activity-conditioned time reparameterization.
Consider a continuous relaxation of the delayed spike arrival map, where an event emitted at time $t'$ arrives at the soma at time $t$ with
\begin{equation}
    t' = t - d(t).
\end{equation}
For a short input window $\Delta t'$ mapped to a somatic integration window $\Delta t$, local differentiation gives
\begin{equation}
    \frac{\Delta t'}{\Delta t} \approx \frac{\partial t'}{\partial t}
    = 1 - d'(t),
    \qquad
    \Delta t \approx \frac{\Delta t'}{1-d'(t)}.
\end{equation}
Thus, in local high-load rising segments where the slope-limited dynamic delay satisfies $0<d'(t)<1$, the effective integration window is elastically dilated.
This dilation spreads spike arrivals across a wider somatic window, reducing instantaneous update pressure rather than merely shifting all spikes by a fixed offset.
In the LIF update, this lowers extreme pre-reset overshoot and keeps membrane potentials closer to the active region of the surrogate derivative, improving temporal credit assignment during BPTT.
Detailed derivations are provided in Appendix~\ref{app:theory}.

\begin{figure}
    \centering
    \includegraphics[width=0.7\linewidth]{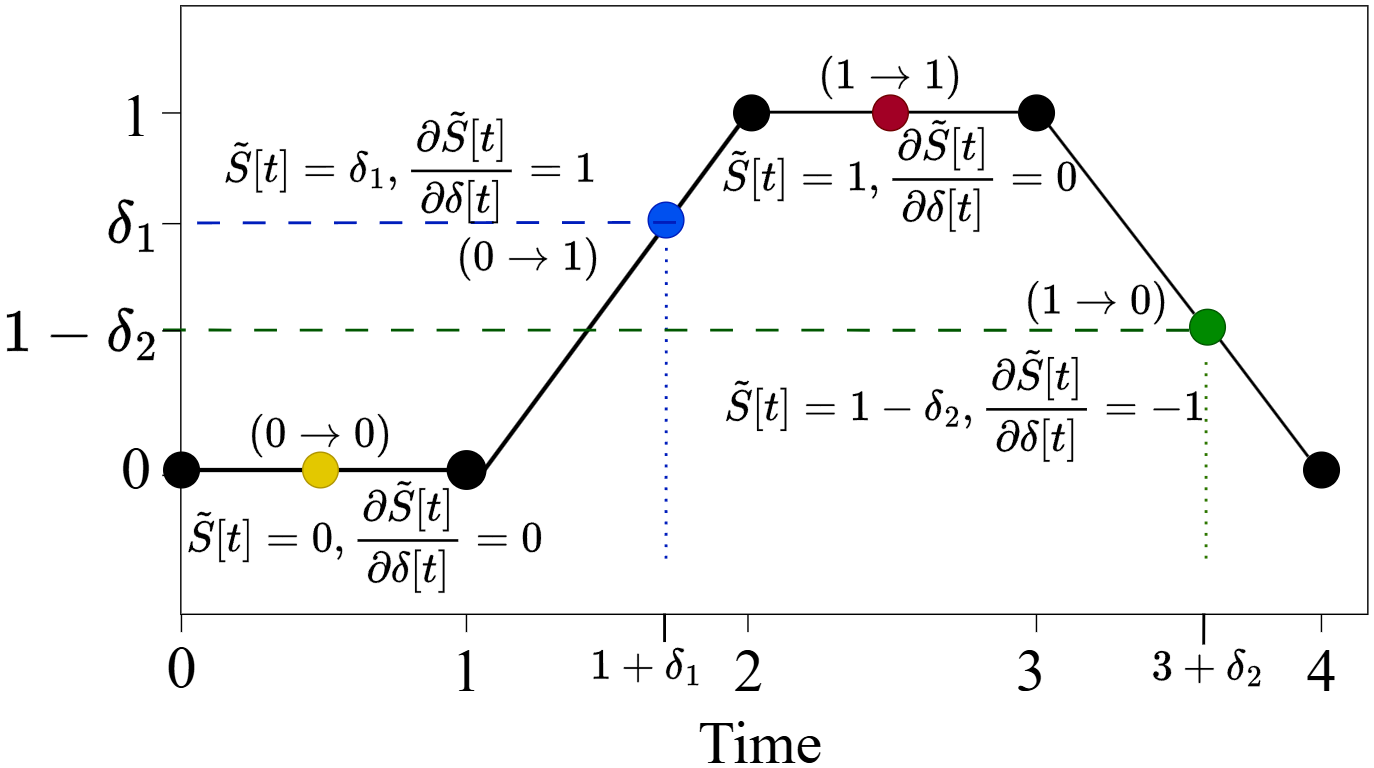}
    \caption{Illustration of differentiable delay approximation via linear interpolation.}
    \label{fig:linear_inter}
\end{figure}

\subsection{Differentiable delay approximation via linear interpolation}

Direct indexing at $t' = t - d[t]$ is undefined on a discrete temporal grid and non-differentiable.
While kernel-based delay modeling has been explored in prior work, such approaches would require generating and applying a distinct convolutional kernel for each time step $t$, resulting in prohibitive computational and memory overhead.
Instead, we approximate the signal value at $t'$ using linear interpolation, which enables fine-grained temporal delay modulation while maintaining low computational complexity.

Specifically, as illustrated in Figure~\ref{fig:linear_inter}, the discretized delayed output $\tilde{S}_i[t]$ is obtained by interpolating between the two nearest integer time steps:
\begin{equation}
    \tilde{S}_i[t] = (1 - \delta[t])\, S_i[\lfloor t' \rfloor] + \delta[t]\, S_i[\lceil t' \rceil],
    \label{interpolation}
\end{equation}
where $\lfloor \cdot \rfloor$ and $\lceil \cdot \rceil$ denote the floor and ceiling operations, respectively, and $\delta[t] = t' - \lfloor t' \rfloor$ represents the fractional temporal offset.
This interpolation scheme supports fine-grained, time-varying delay adaptation with constant computational complexity, while preserving differentiability with respect to both static and dynamic delay parameters during training.
During inference, delays are discretized to integer time steps, ensuring compatibility with standard SNN execution without introducing additional runtime overhead.
This rounding can make adjacent lookup positions equal, but the slope-limited continuous delay prevents temporal order reversal before discretization.


\section{Experiments}
To verify the effectiveness of our method, we conduct experiments on speech benchmarks including SHD (Spiking Heidelberg Digits), SSC (Spiking Speech Commands), and GSC (Google Speech Commands v0.02), all of which require temporal spike-pattern modeling for high classification accuracy. We use a simple feedforward SNN framework and follow the experimental setup of \citet{hammouamri2024learning}. Full hyperparameters are reported in Appendix~\ref{app:sensitivity}.

\begin{table*}[t]
    \centering
    \caption{Classification accuracy on SHD, SSC and GSC-35 datasets. ``Rec.'' denotes recurrent connections. For our methods, values after $\pm$ denote standard deviations across repeated runs; baseline values follow the cited papers.}
        \label{table:results}
    \resizebox{\textwidth}{!}{%
    \begin{tabular}{llccccl}
        \toprule
            Dataset  &  Method & Neuron & Rec. & Delays   & Params  & Top1 Acc.  \\
        \midrule 
            \multirow{11}{4em}{\textbf{SHD}}
            & \small Cuba-LIF  \cite{spikGRU} & Cuba-LIF & \checkmark & \xmark  & 0.14M & 87.80\% \\
            & \small SNN+Delays  \cite{iscas} & LIF & \xmark & \checkmark & 0.1M & 90.43\% \\
            & \small STSC-SNN  \cite{FFSNNattention} & LIF & \xmark & \xmark & 2.1M &  92.36\%      \\
            & \small Adaptive Delays  \cite{sun23} & SRM & \xmark & \checkmark & 0.1M &  92.45\%      \\
            & \small DL128-SNN-Dloss  \cite{sun23-2} & SRM & \xmark & \checkmark & 0.14M &  92.56\%      \\
            & \small RadLIF  \cite{baseline} & ALIF & \checkmark & \xmark    & 3.9M &  94.62\%      \\
            & \small d-cAdLIF \cite{deckersColearningSynapticDelays2024} & ALIF & \xmark & \checkmark & 0.08M &  94.85\% \\
            & \small DCLS-Delays \cite{hammouamri2024learning} & LIF & \xmark & \checkmark & 0.2M &  95.07\% \\
            & \small SE-adLIF \cite{baronig2025advancing} & SE-adLIF & \checkmark & \xmark & 0.45M &  93.79\% \\
            & \small EventProp-Delays \cite{meszaros2025efficient} & LIF & \checkmark & \checkmark & 1.3M &  93.24\% \\
            & \small \textbf{CADAD (2L)}  & \textbf{LIF} & \xmark & \checkmark & \textbf{0.1M} &  \textbf{93.75$\pm$0.80\%} \\
        \midrule
            \multirow{11}{4em}{\textbf{SSC}}
            & \small Heter. RSNN  \cite{heterogeneity}  & ALIF & \checkmark & \xmark & N/A &  57.30\% \\
            & \small SNN-CNN  \cite{ieee_cnn} & LIF & \xmark & \checkmark  & N/A &  72.03\%      \\
            & \small Adaptive SRNN  \cite{Adaptive-SRNN} & ALIF & \checkmark & \xmark  & N/A &  74.20\%      \\
            & \small SpikGRU  \cite{spikGRU} & Cuba-LIF & \checkmark & \xmark  & 0.28M &  77.00\%      \\
            & \small RadLIF  \cite{baseline} & ALIF & \checkmark & \xmark  & 3.9M &  77.40\%      \\
            & \small d-cAdLIF \cite{deckersColearningSynapticDelays2024} & ALIF & \xmark & \checkmark & 0.7M &  80.23\% \\
            & \small DCLS-Delays \cite{hammouamri2024learning} & LIF & \xmark & \checkmark & 1.2M &  80.29\% \\
            & \small SE-adLIF \cite{baronig2025advancing} & SE-adLIF & \checkmark & \xmark & 1.6M &  80.44\% \\
            & \small EventProp-Delays \cite{meszaros2025efficient} & LIF & \xmark & \checkmark & 1.3M &  76.10\% \\
            & \small \textbf{CADAD (2L)}  & \textbf{LIF} & \xmark & \checkmark & \textbf{0.3M} &  \textbf{79.28$\pm$0.11\%} \\
            & \small \textbf{CADAD (3L)}  & \textbf{LIF} & \xmark & \checkmark & \textbf{0.6M} &  \textbf{80.69$\pm$0.24\%} \\
        \midrule
            \multirow{7}{4em}{\textbf{GSC-35}}
            & \small MSAT  \cite{msat} & LIF & \xmark & \xmark & N/A &  87.33\%      \\
            & \small RadLIF  \cite{baseline} & ALIF & \checkmark & \xmark  & 1.2M &  94.51\%      \\
            & \small STSA \cite{wang2023spatial} & LIF & \xmark & \checkmark & N/A &  95.18\% \\
            & \small d-cAdLIF \cite{deckersColearningSynapticDelays2024} & ALIF & \xmark & \checkmark & 0.6M &  95.69\% \\
            & \small DCLS-Delays \cite{hammouamri2024learning} & LIF & \xmark & \checkmark & 1.2M &  95.29\% \\
            & \small \textbf{CADAD (2L)}  & \textbf{LIF} & \xmark & \checkmark & \textbf{0.3M} &  \textbf{95.44$\pm$0.09\%} \\
            & \small \textbf{CADAD (3L)}  & \textbf{LIF} & \xmark & \checkmark & \textbf{0.6M} &  \textbf{95.58$\pm$0.15\%} \\
        \bottomrule
  \end{tabular}
    }
\end{table*}

\begin{table}[t]
\centering
\caption{Controlled ablation study on different delay modeling strategies under the same three-layer feedforward backbone. The CADAD value follows the repeated-run average reported in Table~\ref{table:results}.}
\label{ablation}
\small
\begin{tabular}{lccc}
\toprule
Model & Delay Level & \#Params & Acc. \\
\midrule
No Delay SNN & None  & 0.6M & 59.31\% \\
Dense Conv Delays & Synaptic  & 19M & 78.44\% \\
Static Delays & Axonal & 0.6M & 78.96\% \\
CADAD & Axonal  & 0.6M & 80.69\% \\
\bottomrule
\end{tabular}
\end{table}

\subsection{SHD}
The Spiking Heidelberg Digits (SHD) dataset is a neuromorphic speech benchmark for evaluating SNNs, consisting of approximately 10k spoken digit recordings in English and German converted into multi-channel spike trains by an artificial cochlear front-end. Its rich temporal structure makes SHD particularly suitable for assessing precise spike-timing and delay-aware temporal processing.

We evaluate our method using a feedforward SNN backbone with two hidden layers. Our approach achieves a test accuracy of 93.75$\pm$0.80\% on SHD. Notably, SHD does not provide a dedicated validation split; therefore, consistent with standard practice, we use the test set for validation and early stopping. As discussed in \cite{meszaros2025efficient}, due to the relatively small test set size (2264 samples), accuracy differences beyond a narrow margin are unlikely to be statistically significant. Consequently, although our result is marginally lower than some reported state-of-the-art numbers, it remains well within the expected variance range and provides a fair assessment of model performance.

\subsection{SSC/GSC}
Compared with SHD, the Spiking Speech Commands (SSC) and Google Speech Commands (GSC) datasets provide larger-scale and more challenging speech benchmarks, derived from over 100k one-second spoken utterances. GSC contains raw-audio recordings of 35 common English words, while SSC converts the same recordings into spike trains using an artificial cochlear model. Their greater speaker, pronunciation, and background variability makes them well suited for evaluating temporal robustness under non-stationary spiking activity.

Using a purely feedforward SNN architecture with the same number of hidden layers as competing methods, our approach achieves competitive or better accuracy while using fewer parameters. As reported in Table~\ref{table:results}, our model achieves 80.69$\pm$0.24\% accuracy on SSC and 95.58$\pm$0.15\% accuracy on GSC with only 0.6M parameters. Compared with other works that use synaptic delays, our reparameterized dynamic delay significantly contributes to reducing the number of parameters.

\subsection{Ablation study}
To further assess the effectiveness of the proposed dynamic delay, we conduct ablation experiments on the SSC dataset. We compare four methods on a feedforward SNN with three hidden fully connected layers. The Dense Conv Delays model, introduced in \cite{hammouamri2024learning}, serves as a typical synaptic delay model that enumerates all possible discrete delays as independent synaptic connections. The Static Delays model is the static version of our delay model where $d^{\mathrm{shift}}[t]$ is removed from $d[t]$. The results are shown in Table~\ref{ablation}. Notably, CADAD achieves the best performance in this ablation, demonstrating that input-conditioned, time-varying temporal reparameterization provides additional benefits beyond static temporal offsets. These results suggest that effective delay modeling in SNNs requires structured temporal modulation rather than dense synaptic delay enumeration, and that dynamically adapting delays based on network activity is beneficial for robust temporal processing.

\section{Internal dynamics analysis}
\label{sec:internal_dynamics}
Beyond classification accuracy, we further analyze whether dynamic delays actually reduce the instantaneous state-update pressure suggested by the theoretical interpretation.
We measure the effective input congestion of each layer by
\begin{equation}
    u=\max _{t \in[1, T]}\sum_i |w_i|\,x_i(t),
\end{equation}
which directly reflects the largest delayed input drive entering the LIF update.

\begin{table*}[t]
  \centering
    \caption{Quantitative analysis of input congestion and encoding precision on the SSC dataset. `K' denotes thousands, and `V' represents the unit of membrane potential.}
      \label{tab:comprehensive_comparison}
    \resizebox{\textwidth}{!}{%
    \begin{tabular}{lccccccccc}
    \toprule
    \multirow{2}{*}{\textbf{Layer}} & \multicolumn{2}{c}{\textbf{$u$} (K)} & \multicolumn{2}{c}{\textbf{Overflow} (KV)} & \multicolumn{2}{c}{\textbf{Spikes} (K)} & \multicolumn{2}{c}{\textbf{Overflow per spike} (V)} \\
    \cmidrule(lr){2-3} \cmidrule(lr){4-5} \cmidrule(lr){6-7} \cmidrule(lr){8-9}
          & \textbf{Static} & \textbf{Dynamic} & \textbf{Static} & \textbf{Dynamic} & \textbf{Static} & \textbf{Dynamic} & \textbf{Static} & \textbf{Dynamic} \\
    \midrule
    Layer 0 & 116.5 & \textbf{104.6} & 404.8 & \textbf{379.3} & 659.3 & \textbf{685.9} & 0.614 & \textbf{0.553} \\
    Layer 1 & 100.9 & \textbf{92.2}  & \textbf{535.6} & 599.2 & 953.8 & \textbf{1014.5} & \textbf{0.562} & 0.591 \\
    Layer 2 & 126.9 & \textbf{115.3} & 55.0  & \textbf{48.8}  & \textbf{93.4}  & 88.1  & 0.589 & \textbf{0.554} \\
    \midrule
    \textbf{Sum} & 344.3 & \textbf{312.1} & \textbf{995.4} & 1027.3 & 1706.5 & 1788.6 & 0.583 & \textbf{0.574} \\
    \bottomrule
    \end{tabular}
    }
\end{table*}

As shown in Table~\ref{tab:comprehensive_comparison}, dynamic delays consistently reduce the peak input load $u$ across all hidden layers.
For example, Layer 0 decreases from 116.5K to 104.6K, indicating that the activity-conditioned shift spreads incoming events away from the most congested update moments.
This reduction supports the view that CADAD is not merely adding a static temporal offset, but reshaping the effective integration window under high-load regimes.

We also quantify the information discarded by hard reset using
\begin{equation}
    \mathrm{Overflow}=\sum_t \max(0, v_{\mathrm{inst}}(t)-V_{\mathrm{th}}).
\end{equation}
Although the dynamic model produces more total spikes (1788.6K vs. 1706.5K), it achieves lower overflow per spike (0.574 V vs. 0.583 V).
This means that its spikes are triggered with less residual membrane potential being discarded, which is consistent with sharper threshold crossings and better temporal credit assignment.

\begin{figure*}[t]
  \centering
  \begin{subfigure}[b]{0.48\textwidth}
    \centering
    \includegraphics[width=\linewidth]{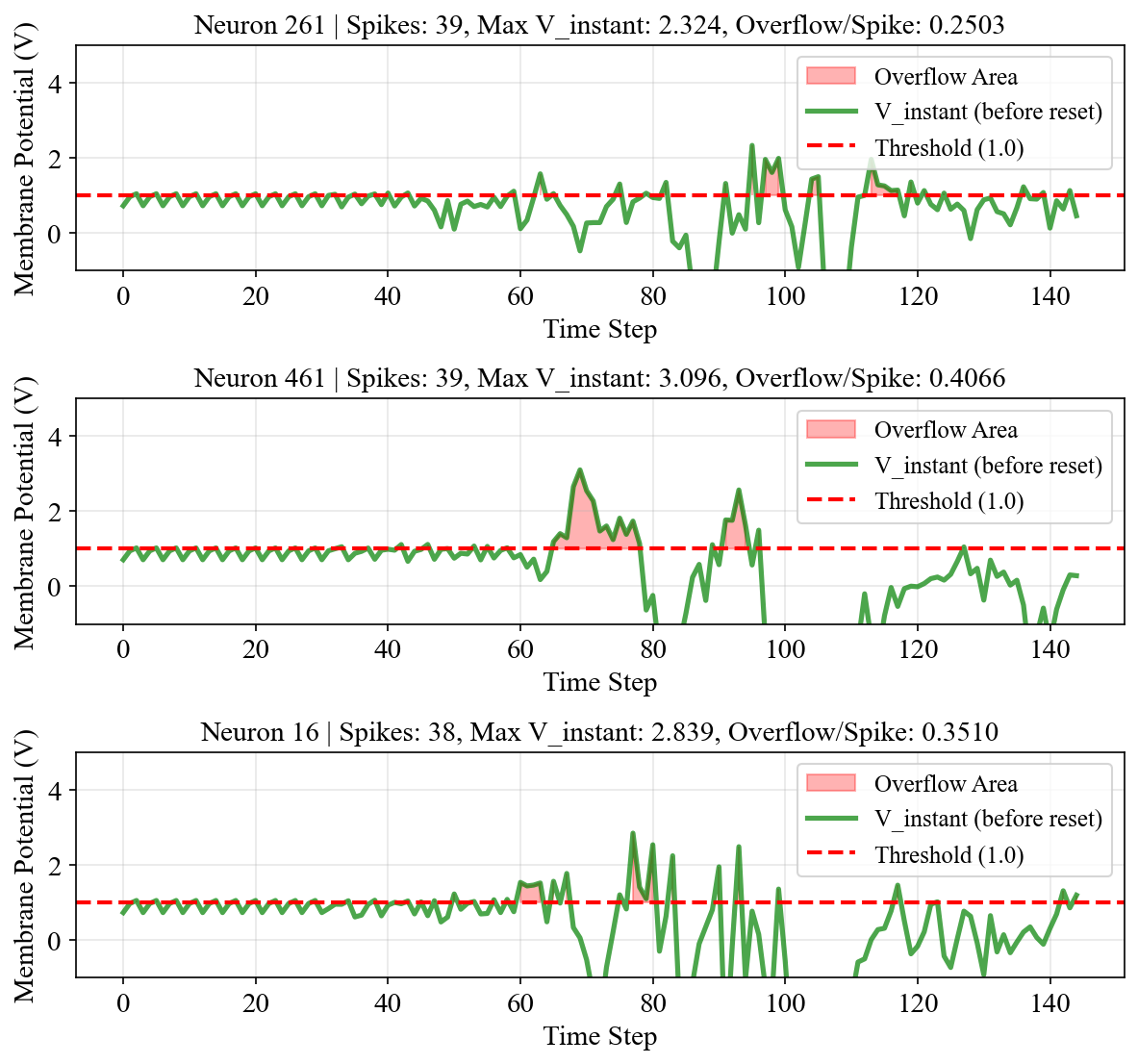}
    \caption{Static delays}
    \label{fig:main_membrane_static}
  \end{subfigure}
  \hfill
  \begin{subfigure}[b]{0.48\textwidth}
    \centering
    \includegraphics[width=\linewidth]{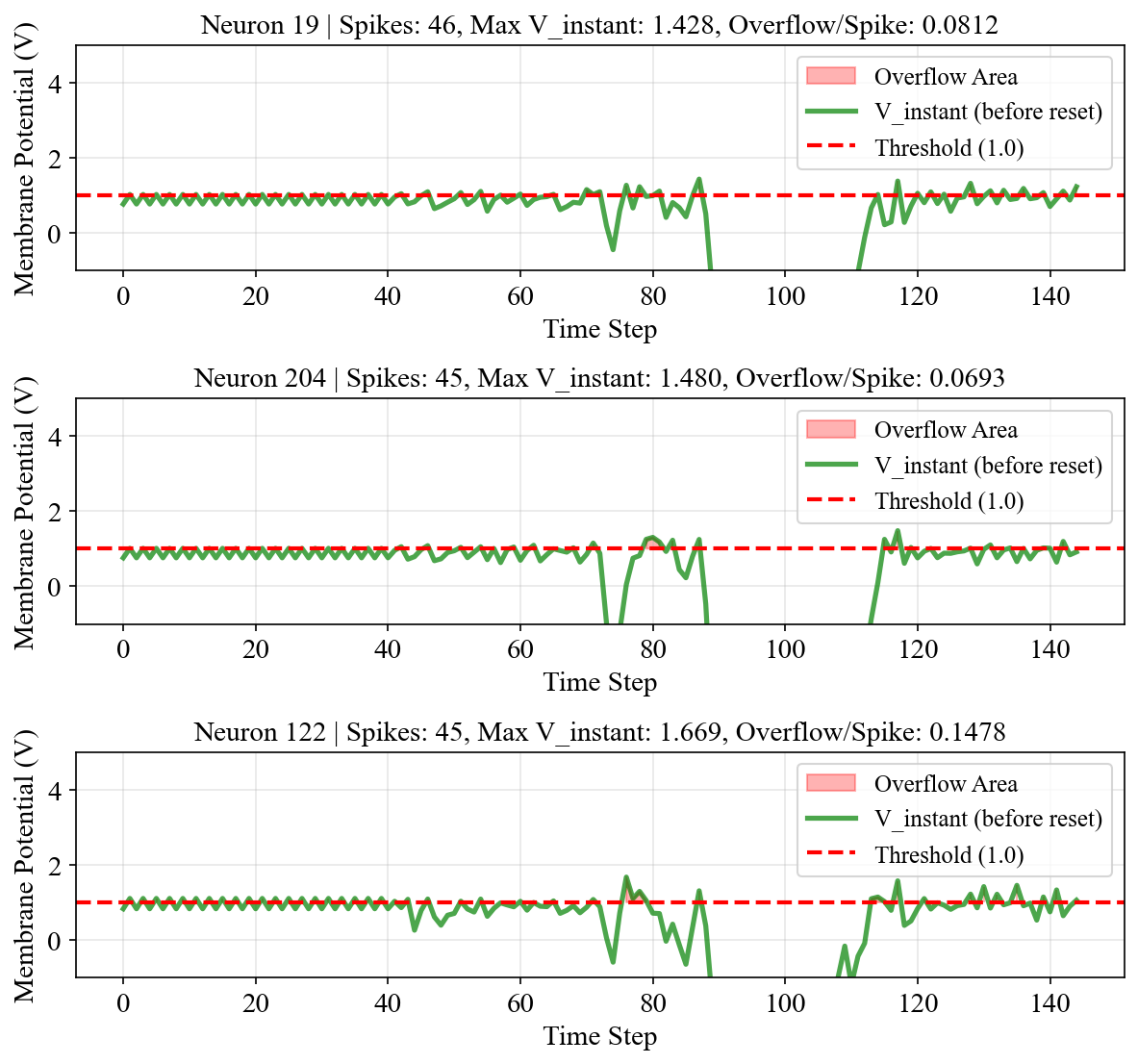} 
    \caption{Dynamic delays}
    \label{fig:main_membrane_dynamic}
  \end{subfigure}
  
  \caption{Visualization of membrane potential traces from the top-3 most active neurons in Layer 0 (Batch 0). The green curves represent the instantaneous potential before reset, and the red shaded regions denote the overflow (discarded magnitude). The static model (a) exhibits ``sluggish'' crossings with large wasteful areas, whereas the dynamic model (b) demonstrates decisive, time-locked firing with high peaks but low average waste, visually corroborating the lower overflow-per-spike metric.}
  \label{fig:membrane_comparison}
\end{figure*}

Figure~\ref{fig:membrane_comparison} visualizes the instantaneous membrane potential traces of representative active neurons in Layer 0, complementing the quantitative internal dynamics analysis in Table~\ref{tab:comprehensive_comparison}.
The green curves represent the theoretical potential before reset, while the red shaded regions denote the overflow area.
In the Static Delay model (Figure~\ref{fig:main_membrane_static}), the membrane potentials exhibit erratic fluctuations with broad, irregular overshoot patterns.
Conversely, the Dynamic Delay model (Figure~\ref{fig:main_membrane_dynamic}) displays significantly sharper and more rhythmic activation patterns.
This cleaner firing behavior demonstrates that dynamic delays align input spikes to trigger instantaneous, high-precision responses, thereby reducing information leakage per spike and improving encoding efficiency.

\section{Conclusion}

We presented a congestion-aware dynamic axonal delay mechanism for spiking neural networks, which reformulates temporal alignment as a signal-level, activity-conditioned reparameterization rather than a static structural property. By decomposing transmission latency into a channel-wise static base delay and a global dynamic shift modulated by network activity, the proposed method dynamically adjusts spike propagation between layers according to the current spike congestion. This design enables fine-grained temporal alignment while preserving event-level semantics, and supports efficient end-to-end optimization through a differentiable linear interpolation scheme.

Experiments on multiple speech benchmarks show that the proposed method improves encoding precision and accuracy while significantly reducing the number of parameters. The current evaluation focuses on temporally driven speech benchmarks, and systematic integration into convolutional or residual neuromorphic vision backbones remains future work. Extending CADAD to broader event-driven time-series tasks and neuromorphic vision backbones is a promising direction for temporally adaptive spiking computation.

\bibliographystyle{plainnat}
\bibliography{neurips26}

\clearpage
\appendix

\section{Theoretical analysis of CADAD}
\label{app:theory}

\subsection{Impact on the neural receptive field}
To analyze how CADAD affects temporal integration, we examine the mapping between a spike's emission time $t'$ and its arrival time $t$:
\begin{equation}
    t' = t - d(t).
\end{equation}
Consider spikes emitted within a short temporal window $\Delta t'$ and integrated over a somatic window $\Delta t$ after the dynamic delay transformation.
Under a local continuous approximation, differentiating with respect to $t$ yields
\begin{equation}
    \frac{\Delta t'}{\Delta t}
    \approx
    \frac{\partial t'}{\partial t}
    =
    1 - d'(t),
    \qquad
    \Delta t
    \approx
    \frac{\Delta t'}{1-d'(t)}.
\end{equation}
In implementation, the dynamic shift is constrained by the discrete slope limiter
\begin{equation}
    |d[t]-d[t-1]| \le 0.99 .
\end{equation}
This ensures that the discrete emission-time map $t-d[t]$ does not fold over time, since
\begin{equation}
    (t-d[t])-((t-1)-d[t-1])
    = 1-(d[t]-d[t-1]) \ge 1-0.99 > 0 .
\end{equation}
When the network enters a high-load regime and the congestion-aware shift increases smoothly over time, this constraint keeps the local condition $0<d'(t)<1$.
The somatic integration window then becomes larger than the input emission window.
The receptive field therefore undergoes an elastic temporal dilation, allowing the neuron to process incoming spikes over an extended support rather than compressing them into a narrow update interval.

\subsection{Membrane potential stability}
The pre-spike membrane update in the discrete LIF model is
\begin{equation}
    \tilde{v}[t] = \lambda v[t-1] + I[t],
\end{equation}
where $I[t]$ is determined by the delayed incoming spikes.
For static or fixed-shift delays, $d'(t)=0$ and hence $\Delta t=\Delta t'$.
Such delays can align spikes, but they cannot adaptively stretch the arrival window when many spikes are concentrated in time.
In high-load regimes this can produce large instantaneous input currents, causing excessive pre-reset membrane overshoot.

By contrast, the dynamic shift in CADAD acts as a nonlinear time-stretching operator.
When congestion increases under the slope-limited condition $0<d'(t)<1$, the effective integration window is dilated ($\Delta t>\Delta t'$), which distributes incoming spikes across a wider interval.
This structurally reduces instantaneous arrival density and helps clamp the variance of the pre-reset membrane potential.
The effect is not merely a uniform translation of spike trains; it is an activity-dependent redistribution of update pressure.

\subsection{Temporal credit assignment}
Temporal credit assignment in surrogate-gradient SNN training depends on the surrogate derivative $H'(\tilde{v}[t]-V_{\mathrm{th}})$.
Severe congestion may push the membrane far above threshold, $\tilde{v}[t]\gg V_{\mathrm{th}}$, outside the narrow active region of $H'$.
In that case, the corresponding temporal Jacobian terms approach zero, truncating error signals during BPTT.
By reducing extreme overshoot and keeping membrane trajectories closer to threshold-crossing regions, CADAD preserves larger surrogate gradients over time.
This provides a theoretical explanation for why dynamic delay modulation can improve both stability and long-range temporal credit assignment.

\section{Gradient Analysis of Linear Interpolation for Differentiable Delay}
\label{app:gradient_interpolation}
To update the learnable delay $d[t]$ using gradient descent, we compute the partial derivative of the loss $L$ with respect to $d[t]$. Applying the chain rule:
\begin{equation}
    \frac{\partial L}{\partial d[t]} = \frac{\partial L}{\partial \tilde{S}_i[t]} \cdot \frac{\partial \tilde{S}_i[t]}{\partial d[t]},
\end{equation}
where $\frac{\partial L}{\partial \tilde{S}_i[t]}$ is the error gradient propagated from the subsequent layers. For $\frac{\partial \tilde{S}_i[t]}{\partial d[t]}$, the term can be decomposed as
\begin{equation}
    \frac{\partial \tilde{S}_i[t]}{\partial d[t]} = \frac{\partial \tilde{S}_i[t]}{\partial \delta} \cdot \frac{\partial \delta}{\partial t'} \cdot \frac{\partial t'}{\partial d[t]}.
\end{equation}
First, the derivative of interpolation with respect to $\delta$ is
\begin{equation}
\begin{aligned}
        \frac{\partial \tilde{S}_i[t]}{\partial \delta} &= \frac{\partial}{\partial \delta} \left[ (1 - \delta) S_i[t_{\mathrm{floor}}] + \delta S_i[t_{\mathrm{ceil}}] \right]\\
        &= - S_i[t_{\mathrm{floor}}] + S_i[t_{\mathrm{ceil}}] \\
        &= S_i[t_{\mathrm{ceil}}] - S_i[t_{\mathrm{floor}}].
\end{aligned}
\end{equation}
This represents the local slope of the signal between the two discrete time steps. Second, since $\delta = t' - \lfloor t' \rfloor$ and $\lfloor t' \rfloor$ is piecewise constant, the derivative of $\delta$ with respect to $t'$ equals 1 almost everywhere:
\begin{equation}
    \frac{\partial \delta}{\partial t'} = 1.
\end{equation}
Last, for the derivative of $t'$ with respect to $d[t]$:
\begin{equation}
\begin{aligned}
        \frac{\partial t'}{\partial d[t]} &= \frac{\partial (t - d[t])}{\partial d[t]}  \\
        &= -1.
\end{aligned}
\end{equation}
Together:
\begin{equation}
    \frac{\partial \tilde{S}_i[t]}{\partial d[t]} = S_i[t_{\mathrm{floor}}] - S_i[t_{\mathrm{ceil}}],
\end{equation}
\begin{equation}
\begin{split}
    \frac{\partial L}{\partial d[t]} & = \frac{\partial L}{\partial \tilde{S}_i[t]} \cdot (S_i[t_{\mathrm{floor}}] - S_i[t_{\mathrm{ceil}}]) \\
    & = \frac{\partial L}{\partial \tilde{S}_i[t]} \cdot 
    \begin{cases} 
    1, & \text{if } S_i[t_{\mathrm{floor}}]=1 \land S_i[t_{\mathrm{ceil}}]=0 \\
    -1, & \text{if } S_i[t_{\mathrm{floor}}]=0 \land S_i[t_{\mathrm{ceil}}]=1 \\
    0, & \text{otherwise}.
    \end{cases}
\end{split}
\end{equation}
This derivation demonstrates that the gradient for the delay parameter depends on the local temporal difference of the input signal. Importantly, the gradient with respect to the delay parameter is proportional to the local temporal difference of the input signal, enabling the delay to adaptively respond to temporal signal dynamics while preserving stable and efficient gradient propagation.

\section{Ablation on the Non-linear Function Applied to Generating Dynamic Shift}
\label{app:nonlinear_ablation}
To investigate the impact of different non-linear mapping functions on the dynamic delay generation, we conducted an ablation study by replacing the hyperbolic tangent (Tanh) in Eq.~\ref{eq:tanh} with other common activation functions: Sigmoid, ReLU, and Arctan.
The results on the SSC dataset are reported in Table~\ref{tab:act}.
The results indicate that bounded S-shaped functions like Tanh and Sigmoid generally outperform the unbounded function ReLU, as unbounded delays under extreme congestion might lead to excessive temporal shifts, potentially disrupting the causal structure of event sequences.
Among the S-shaped curves, Arctan performs worse.
We attribute this to the gradient characteristics near zero: Tanh has a steeper derivative at the origin compared to Arctan, allowing the mechanism to react more promptly to the onset of input congestion.
Sigmoid performs very closely to Tanh, but it would introduce a constant temporal offset even when the network is quiet because Sigmoid(0)=0.5.
Therefore, we use Tanh as the default choice.

\begin{table}[H]
\centering
\begin{tabular}{lc} 
\toprule
Activation  & Acc. \\
\midrule
Tanh  & 80.69\% \\
Sigmoid   & 80.63\% \\
ReLU  & 80.37\% \\
Arctan & 80.23\% \\
\bottomrule
\end{tabular}
\caption{Performance of different non-linear functions on SSC.}
\label{tab:act}
\end{table}

\section{Hyperparameter Sensitivity}
\label{app:sensitivity}

Across SHD, SSC, and GSC, CADAD remains stable under one-factor-at-a-time hyperparameter variations of the activity sensitivity $\gamma$, temporal smoothing window $K_{\mathrm{smooth}}$, dynamic delay range $(S_{\max}, S_{\min})$, and decay duration $E_{\mathrm{decay}}$.
This stability suggests that the gain does not come from a narrow hyperparameter choice, but from the congestion-aware redistribution of update pressure.
This behavior is consistent with the intended factorization: channel-wise base delays provide persistent temporal alignment, while the shared activity-conditioned shift modulates transient update pressure without introducing synapse-specific delay parameters.

\begin{table}[ht]
\centering
\caption{Network parameters used as the default configuration for sensitivity analysis. During each sensitivity experiment, only one hyperparameter is varied while the others are fixed according to this table.}
\label{tab:sensitivity_default_params}
\begin{tabular}{lccc}
\toprule
Dataset & SHD & SSC & GSC \\
\midrule
epochs & 150 & 60 & 100 \\
Batch size & 256 & 128 & 128 \\
Optimizer & Adam & Adam & Adam \\
Weight decay & $1\mathrm{e}{-5}$ & $1\mathrm{e}{-5}$ & $1\mathrm{e}{-5}$ \\
Weight scheduler & OneCycle & OneCycle & OneCycle \\
Delay scheduler & Cosine & Cosine & Cosine \\
$\gamma$ & 1.0 & 1.0 & 1.0 \\
$K_{\mathrm{smooth}}$ & 20 & 20 & 20 \\
$S_{\max}$ & 0.5 & 0.3 & 0.2 \\
$S_{\min}$ & 0.10 & 0.02 & 0.01 \\
$E_{\mathrm{decay}}$ & 70 & 30 & 50 \\
$l r_w$ & $1\mathrm{e}{-3}$ & $1\mathrm{e}{-3}$ & $1\mathrm{e}{-3}$ \\
$l r_{\mathrm{delay}}$ & $1\mathrm{e}{-1}$ & $1\mathrm{e}{-1}$ & $1\mathrm{e}{-1}$ \\
Hidden Layers & 2 & 3 & 3 \\
Hidden size & 256 & 512 & 512 \\
$\tau$ (ms) & 10.05 & 15 & 15 \\
Maximum Delay (ms) & 250 & 300 & 300 \\
Dropout rate & 0.4 & 0.25 & 0.25 \\
\bottomrule
\end{tabular}
\end{table}

\begin{table*}[t]
\centering
\caption{Hyperparameter sensitivity results on SHD, SSC, and GSC (all results are obtained using seed 0). The reported values are best accuracy (\%). For SSC and GSC, test accuracy is reported. For SHD, the reported value is the best validation accuracy, following the original data split used in our implementation.}
\label{tab:sensitivity_results}
\begin{tabular}{llccc}
\toprule
Hyperparameter & Value & SHD & SSC & GSC \\
\midrule
\multirow{5}{*}{$\gamma$}
& 0.25 & 93.93 & 80.41 & 95.42 \\
& 0.5  & 93.72 & 80.52 & 95.40 \\
& 1.0  & 93.75 & 80.69 & 95.58 \\
& 1.5  & \textbf{94.61} & 80.50 & \textbf{95.61} \\
& 2.0  & 94.10 & \textbf{81.04} & 95.38 \\
\midrule
\multirow{5}{*}{$K_{\mathrm{smooth}}$}
& 5  & 94.19 & 80.53 & \textbf{95.60} \\
& 10 & 94.32 & 79.97 & 95.58 \\
& 20 & 93.75 & 80.69 & 95.58 \\
& 40 & 94.00 & \textbf{80.77} & 95.48 \\
& 80 & \textbf{94.47} & 80.27 & 95.47 \\
\midrule
\multirow{5}{*}{$S_{\max}$}
& 0.1 & 94.10 & 80.65 & 95.47 \\
& 0.2 & \textbf{94.18} & \textbf{80.92} & \textbf{95.58} \\
& 0.3 & 93.82 & 80.69 & 95.20 \\
& 0.4 & 93.99 & 80.55 & 95.14 \\
& 0.5 & 93.75 & 80.81 & 95.15 \\
\midrule
\multirow{5}{*}{$S_{\min}$}
& 0.0  & \textbf{94.56} & 79.36 & \textbf{95.63} \\
& 0.01 & 93.58 & 80.54 & 95.58 \\
& 0.02 & 93.75 & 80.69 & 95.55 \\
& 0.05 & 94.00 & \textbf{80.79} & 95.47 \\
& 0.1  & 93.75 & 80.13 & 95.41 \\
\midrule
\multirow{5}{*}{$E_{\mathrm{decay}}$}
& 0    & 91.68 & 79.82 & 95.25 \\
& low  & 93.56 & 80.50 & \textbf{95.67} \\
& mid  & 93.67 & \textbf{80.74} & 95.36 \\
& default & 93.75 & 80.69 & 95.58 \\
& full & \textbf{94.25} & 80.64 & 95.54 \\
\bottomrule
\end{tabular}
\end{table*}

\begin{table}[ht]
\centering
\caption{Dataset-specific values used for $E_{\mathrm{decay}}$ sensitivity analysis.}
\label{tab:edecay_values}
\begin{tabular}{lccc}
\toprule
Setting & SHD & SSC & GSC \\
\midrule
0 & 0 & 0 & 0 \\
low & 19 & 8 & 13 \\
mid & 38 & 15 & 25 \\
default & 70 & 30 & 50 \\
full & 150 & 60 & 100 \\
\bottomrule
\end{tabular}
\end{table}

We conduct a one-factor-at-a-time hyperparameter sensitivity analysis of the proposed dynamic-delay mechanism. For each dataset, we start from the default configuration in Table~\ref{tab:sensitivity_default_params} and vary only one hyperparameter while keeping all others fixed. The analyzed hyperparameters include the activity sensitivity $\gamma$, the temporal smoothing window $K_{\mathrm{smooth}}$, the maximum and minimum dynamic delay scales $S_{\max}$ and $S_{\min}$, and the decay duration $E_{\mathrm{decay}}$. All experiments are conducted using a fixed seed 0 to ensure consistency across comparisons.

As shown in Table~\ref{tab:sensitivity_results}, the model is generally stable across a broad range of settings. On SHD, the best result is obtained with $\gamma=1.5$, reaching 94.61\%, while several other configurations also remain above 94\%. This suggests that SHD benefits from moderately stronger activity-dependent delay modulation. The performance is less favorable when $E_{\mathrm{decay}}=0$, indicating that immediately suppressing dynamic delay adaptation can hurt temporal representation learning.

For SSC, the best test accuracy is achieved with $\gamma=2.0$, giving 81.04\%. The model is also relatively insensitive to $S_{\max}$ within the tested range, with $S_{\max}=0.2$ yielding 80.92\%. In contrast, setting $S_{\min}=0$ leads to a clear degradation, suggesting that preserving a non-zero lower bound for dynamic delay modulation is important for maintaining useful temporal flexibility throughout training.

On GSC, the overall variation is small, and most configurations achieve test accuracy around 95.4--95.7\%. The best test accuracy is obtained with a shorter decay schedule, $E_{\mathrm{decay}}=13$ (low), reaching 95.67\%. This indicates that GSC may require less prolonged dynamic delay adaptation than SHD, likely because the input representation is more regular and less sparse than event-based speech datasets.

Overall, the hyperparameter sensitivity analysis shows that the proposed dynamic-delay mechanism remains stable under moderate hyperparameter changes. The default settings provide competitive performance across all datasets, while dataset-specific tuning of $\gamma$, $S_{\min}$, and $E_{\mathrm{decay}}$ can further improve accuracy.

\section{Discussion}
\label{app:discussion}

The theoretical view above suggests that CADAD improves temporal processing by reshaping the effective integration window, not simply by adding more delay parameters.
This interpretation is consistent with the internal diagnostics in Section~\ref{sec:internal_dynamics}: dynamic delays reduce peak input load while maintaining strong spike activity and lowering overflow per spike.
Our main experiments follow prior delay-learning studies that use feedforward SNNs to expose temporal alignment effects directly.
Broader neuromorphic vision settings introduce additional factors such as spatial feature extractors, residual blocks, and event-camera preprocessing, which can obscure whether gains arise from the delay mechanism itself or from architecture-specific design.
Since CADAD operates on spike arrival times before the synaptic update, it can in principle be inserted into convolutional or residual SNN layers without changing the surrounding classifier structure.
Systematically integrating CADAD into such backbones is therefore an important direction for evaluating how congestion-aware temporal modulation interacts with spatial hierarchy.

\section{Computational resources}
\label{app:compute}

All reported speech experiments were trained on NVIDIA A100-PCIE GPUs with 40GB memory under CUDA 12.6.
Each individual training run used a single GPU; multiple GPUs were used only to launch independent seeds, ablations, or sensitivity configurations in parallel.
No model-parallel training was used.

The typical wall-clock time per run was approximately 20--30 minutes for SHD with 150 epochs, 1.4--1.9 hours for SSC with 60 epochs, and 3.0--3.3 hours for GSC with 100 epochs.
These estimates include training and validation/testing performed during training, and correspond to the default settings in Table~\ref{tab:sensitivity_default_params}.
The total compute for the main repeated-run results scales linearly with the number of independent seeds; the additional sensitivity analysis in Table~\ref{tab:sensitivity_results} was run as separate one-factor-at-a-time jobs using the same single-GPU setting.

\section{Supplementary Figures of Membrane Potential Traces}
\label{app:supplementary_membrane}
To further substantiate the qualitative analysis presented in Section~\ref{sec:internal_dynamics} regarding encoding precision, we provide additional visualizations of instantaneous membrane potential traces, including neurons from deeper network layers on the SSC dataset.

\begin{figure*}[htbp]
  \centering
  \begin{subfigure}[b]{0.48\textwidth}
    \centering
    \includegraphics[width=\linewidth]{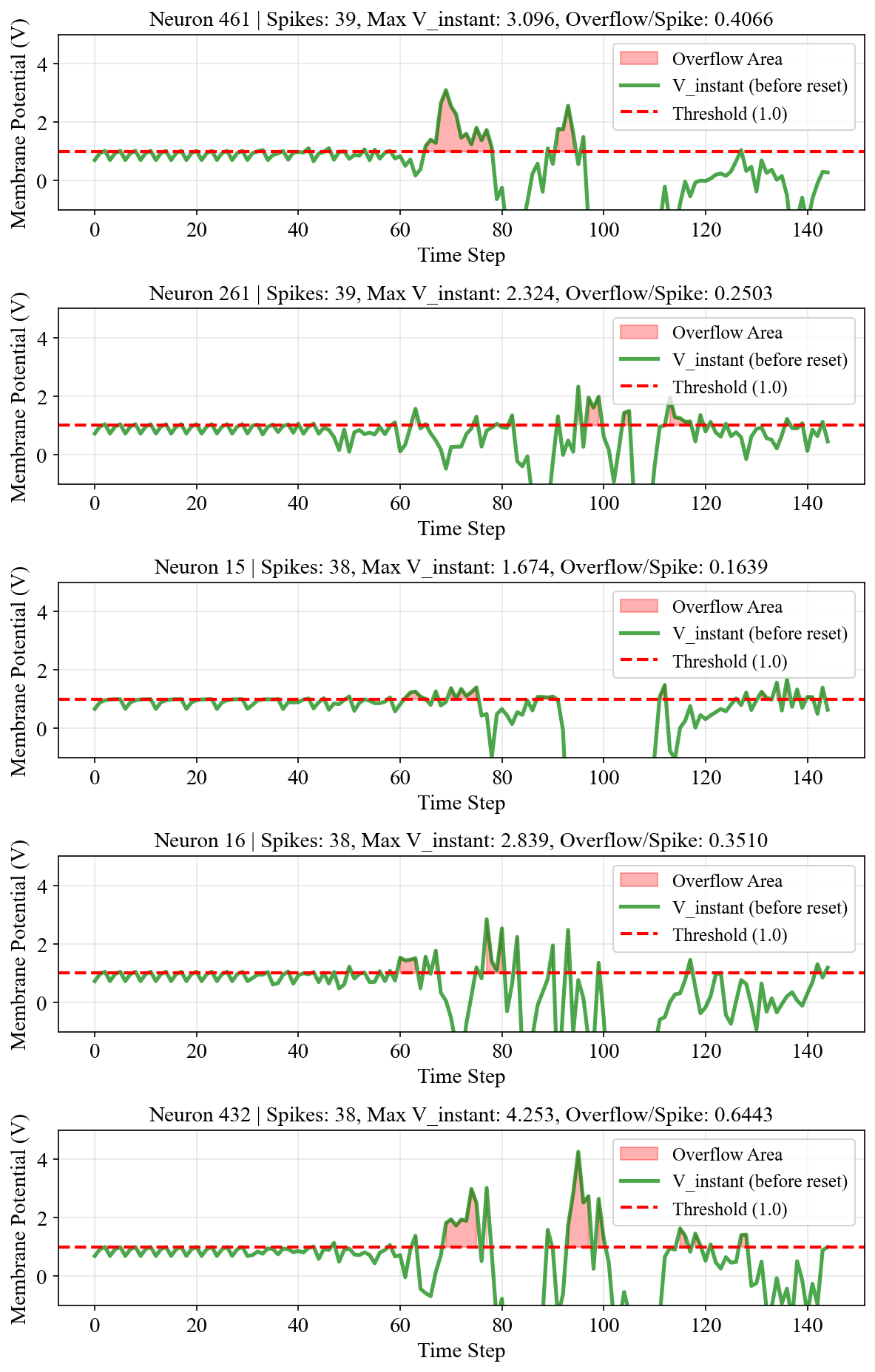}
    \caption{Static delays}
    \label{fig:app_layer0_static}
  \end{subfigure}
  \hfill
  \begin{subfigure}[b]{0.48\textwidth}
    \centering
    \includegraphics[width=\linewidth]{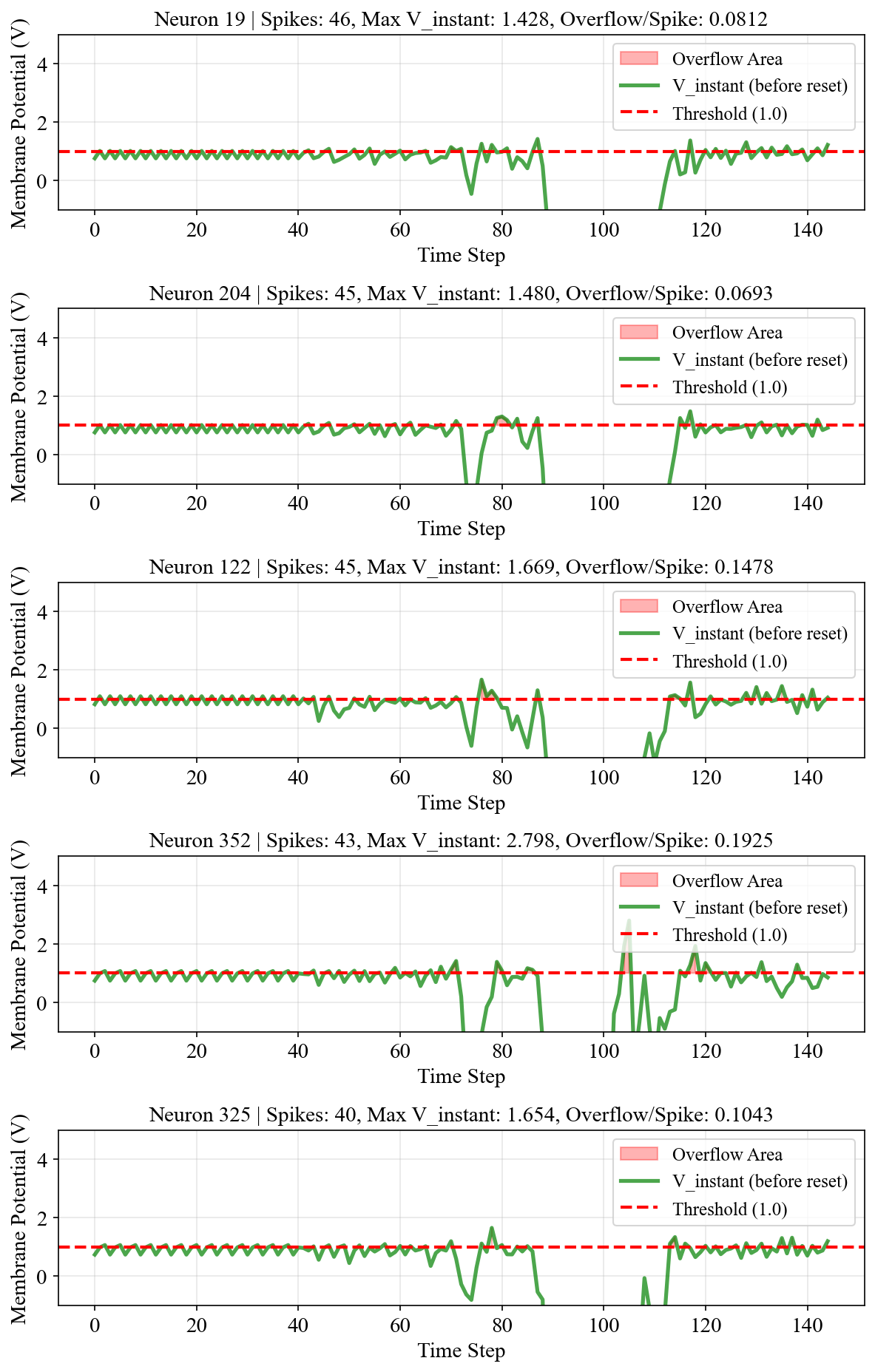} 
    \caption{Dynamic delays}
    \label{fig:app_layer0_dynamic}
  \end{subfigure}
  
  \caption{Visualization of membrane potential traces from the top-5 most active neurons in Layer 0 (Batch 0).}
  \label{fig:app_layer0_membrane}
\end{figure*}

As the immediate interface for input processing, Layer 0 typically experiences the most severe congestion.
In Figure~\ref{fig:app_layer0_membrane}, the effect of our mechanism is most pronounced.

\begin{figure*}[htbp]
  \centering
  \begin{subfigure}[b]{0.48\textwidth}
    \centering
    \includegraphics[width=\linewidth]{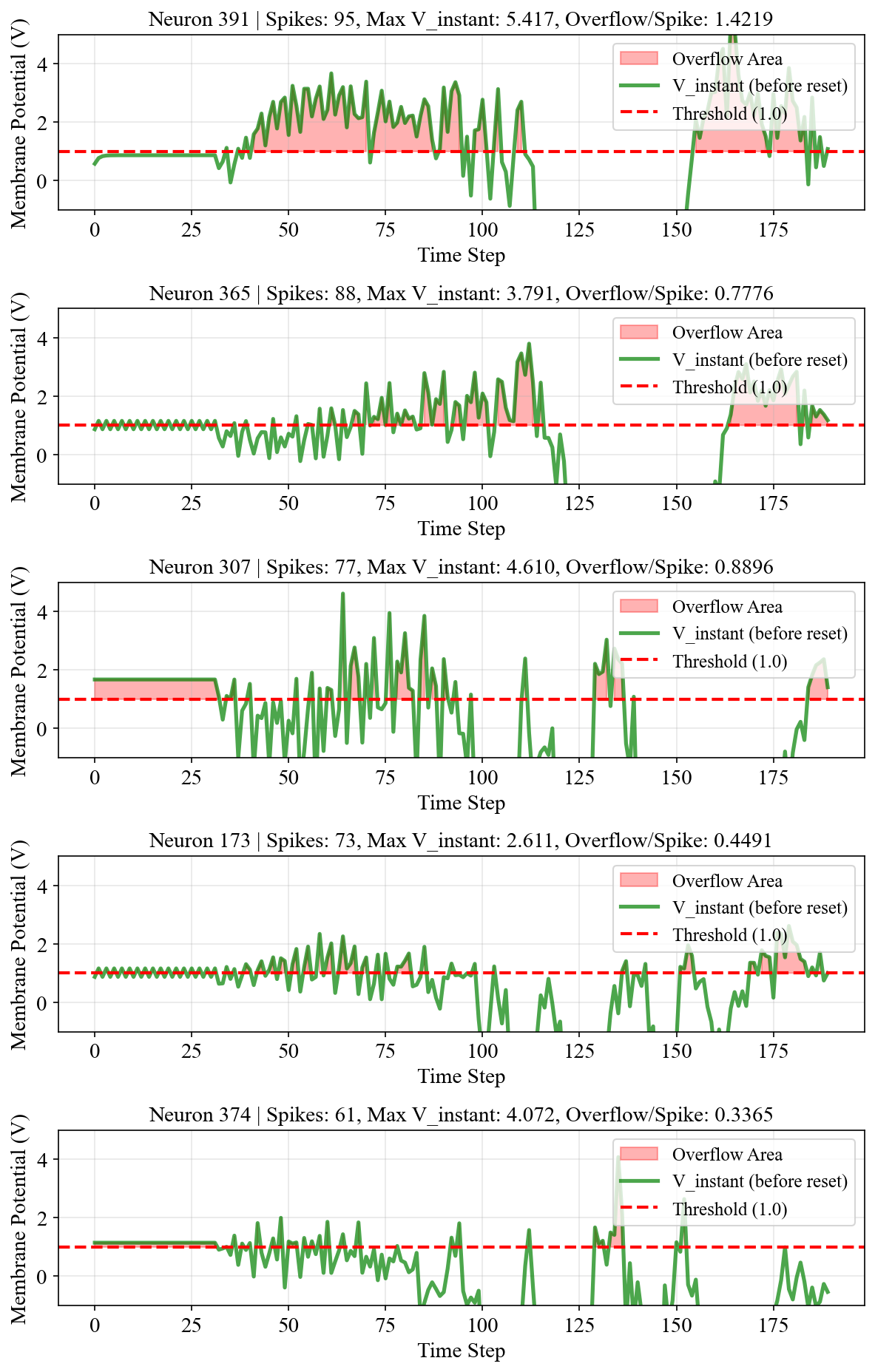}
    \caption{Static delays}
    \label{fig:app_layer1_static}
  \end{subfigure}
  \hfill
  \begin{subfigure}[b]{0.48\textwidth}
    \centering
    \includegraphics[width=\linewidth]{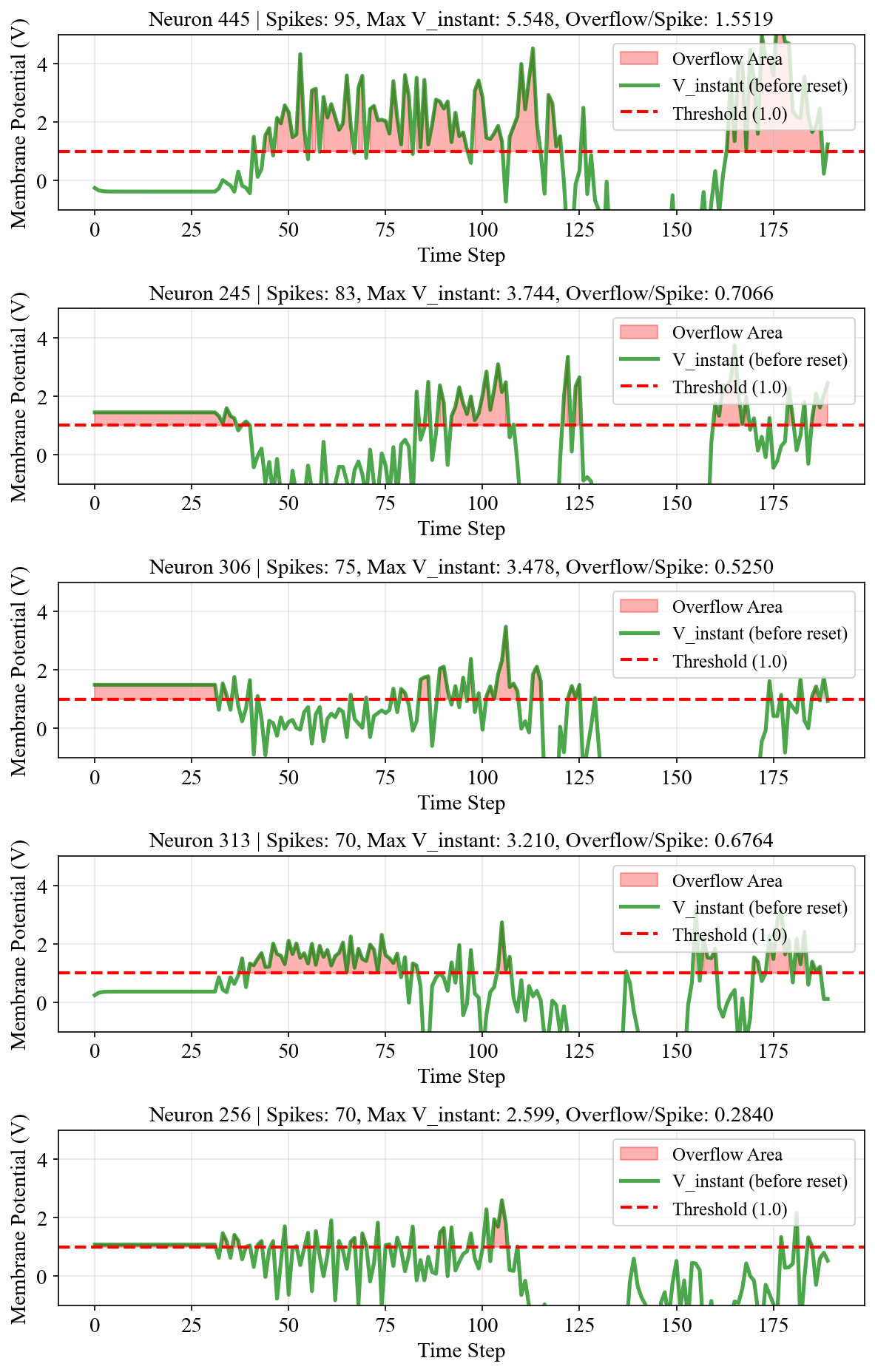} 
    \caption{Dynamic delays}
    \label{fig:app_layer1_dynamic}
  \end{subfigure}
  
  \caption{Visualization of membrane potential traces from the top-5 most active neurons in Layer 1 (Batch 0).}
  \label{fig:app_layer1_membrane}
\end{figure*}

Layer 1 represents the most active stage of the network, characterized by high firing rates.
As shown in Figure~\ref{fig:app_layer1_membrane}, severe saturation-like behavior is observed in the static model.
In this state, the neuron loses sensitivity to specific spike timing, effectively degenerating into a crude rate-coding unit.
In the dynamic model, while the firing rate remains high, dynamic delays force the membrane potential into a high-frequency oscillation pattern.
By dynamically adjusting delays, the mechanism prevents the potential from sticking above the threshold, compelling the neuron to reset and re-integrate inputs.
This preserves the fine-grained temporal structure of the features even under high-intensity conditions.

\begin{figure*}[htbp]
  \centering
  \begin{subfigure}[b]{0.48\textwidth}
    \centering
    \includegraphics[width=\linewidth]{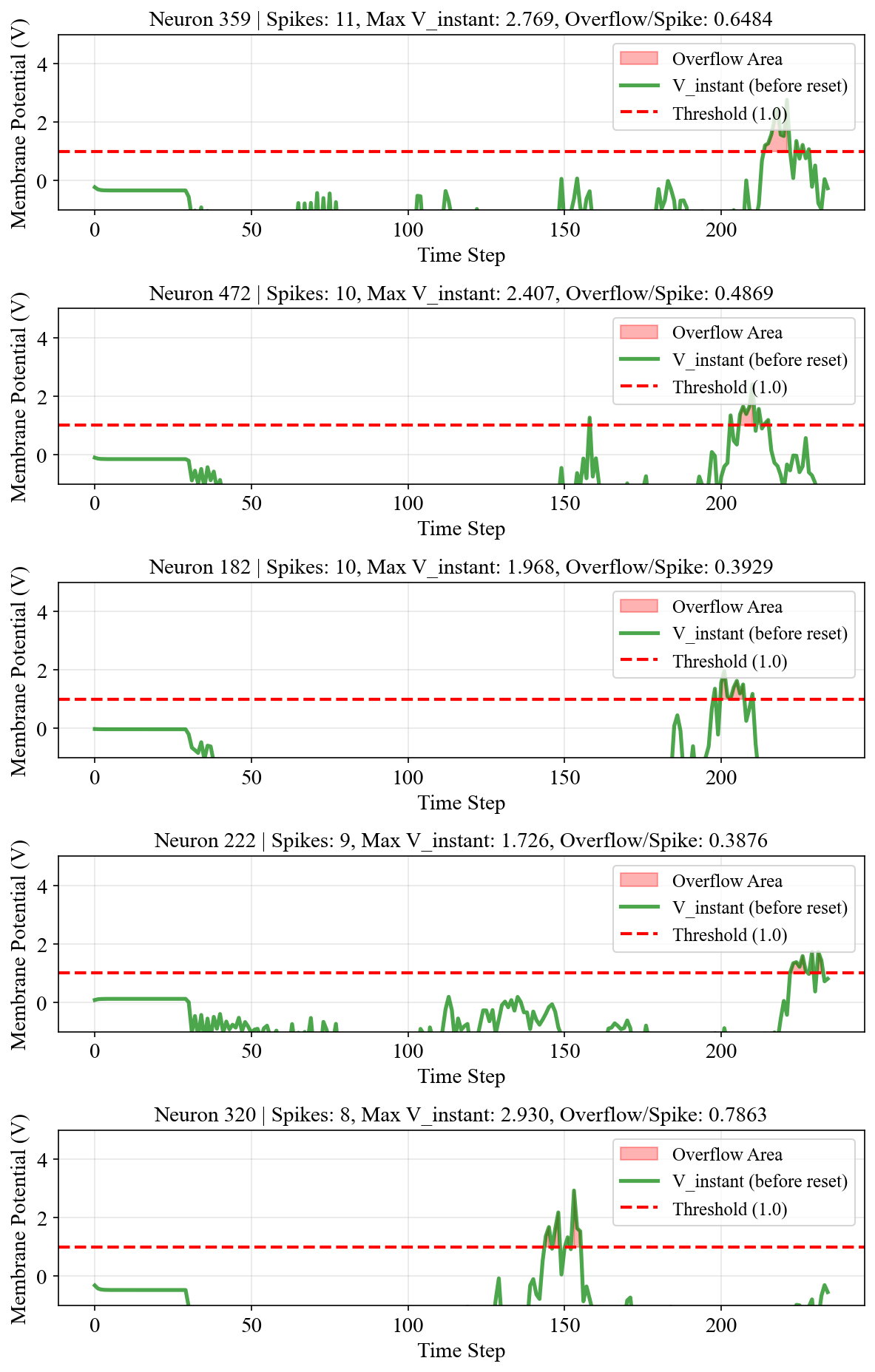}
    \caption{Static delays}
    \label{fig:app_layer2_static}
  \end{subfigure}
  \hfill
  \begin{subfigure}[b]{0.48\textwidth}
    \centering
    \includegraphics[width=\linewidth]{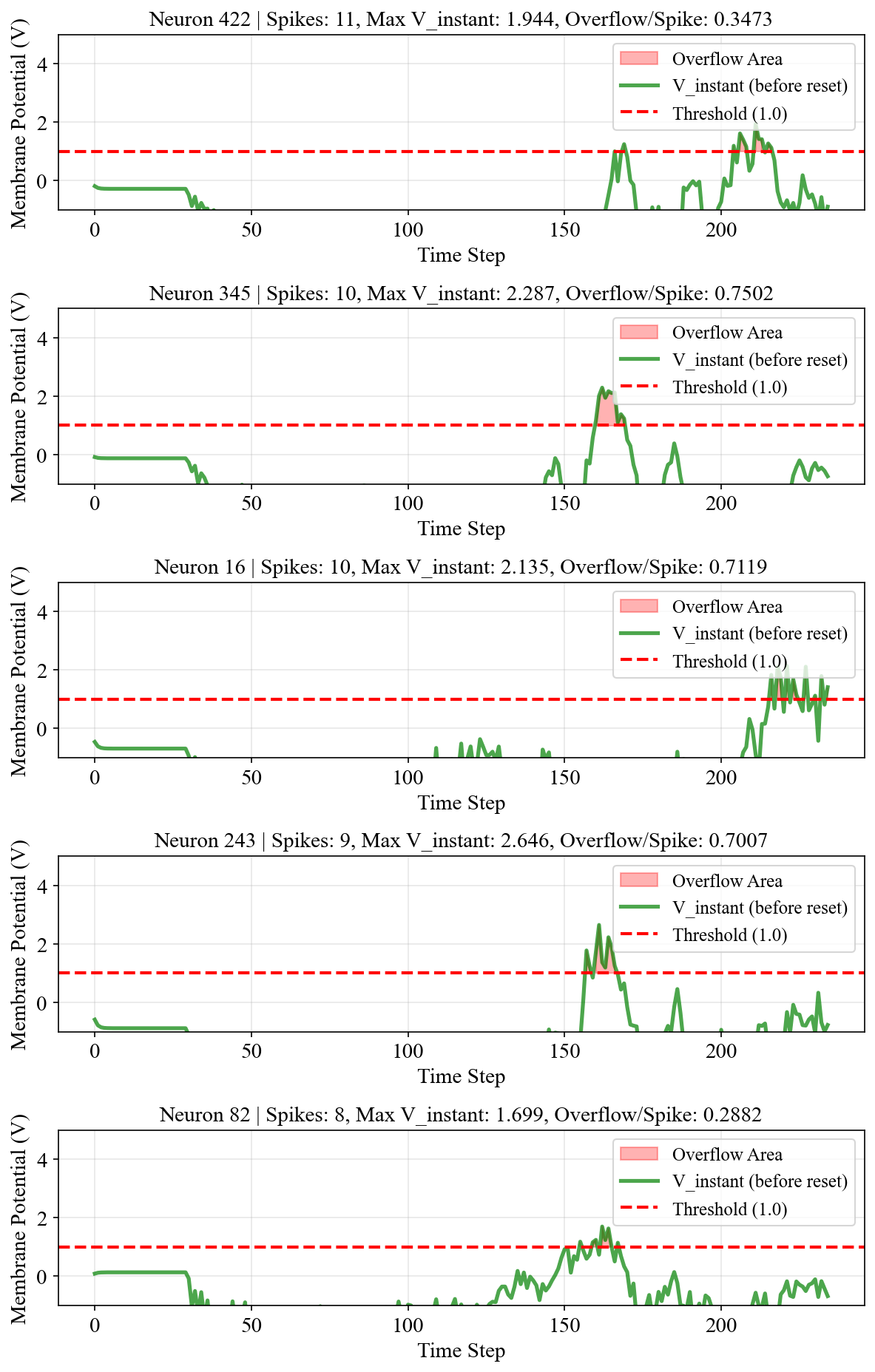} 
    \caption{Dynamic delays}
    \label{fig:app_layer2_dynamic}
  \end{subfigure}
  
  \caption{Visualization of membrane potential traces from the top-5 most active neurons in Layer 2 (Batch 0).}
  \label{fig:app_layer2_membrane}
\end{figure*}

In Layer 2, signals become more sparse (Figure~\ref{fig:app_layer2_membrane}).
Instability persists in the static model, while the dynamic model maintains rapid and sharp responses with lower overflow per spike.

\clearpage

\end{document}